\documentclass[12pt]{article}

\usepackage[utf8]{inputenc}
\usepackage[T1]{fontenc}
\usepackage{lmodern}

\usepackage{amsmath,amssymb,amsfonts,amsthm,bm,bbm,eucal,relsize,latexsym}

\usepackage{booktabs,multirow,makecell,arydshln,longtable}
\usepackage{authblk}
\usepackage{graphicx,adjustbox,subfigure,rotating,lscape,wrapfig,tikz,epstopdf}

\usepackage{algorithm,algpseudocode}

\usepackage{enumerate,enumitem,listings,listofitems,stackengine}

\usepackage{setspace,sectsty,textcomp,textcmds,fancyhdr,verbatim}

\usepackage{natbib,url}

\usepackage{xcolor}
\definecolor{lb}{RGB}{44,139,183}

\usepackage{pdfpages}

\usepackage[colorlinks,citecolor=blue,urlcolor=blue]{hyperref}

\allowdisplaybreaks

\title{Spatio-temporal DeepKriging in PyTorch: A
Supplementary Application to Precipitation Data for
Interpolation and Probabilistic Forecasting}

\author[1]{Pratik Nag}

\affil[1]{School of Mathematics and Applied Statistics, University of Wollongong, Australia}

\date{}

\begin{document}


\maketitle

\begin{abstract}
A detailed analysis of precipitation data over Europe is presented, with a focus on interpolation and forecasting applications. A Spatio-temporal DeepKriging (STDK) framework has been implemented using the PyTorch platform to achieve these objectives. The proposed model is capable of handling spatio-temporal irregularities while generating high-resolution interpolations and multi-step forecasts. Reproducible code modules have been developed as standalone PyTorch implementations for the interpolation\footnote[2]{Interpolation - https://github.com/pratiknag/Spatio-temporalDeepKriging-Pytorch.git} and forecasting\footnote[3]{Forecasting - https://github.com/pratiknag/pytorch-convlstm.git}, facilitating broader application to similar climate datasets. The effectiveness of this approach is demonstrated through extensive evaluation on daily precipitation measurements, highlighting predictive performance and robustness.
\end{abstract}


\baselineskip=26pt

\section{Introduction}

Precipitation plays a critical role in the global hydrological cycle and affects diverse areas such as water resource management, agriculture, disaster preparedness, and climate change research \citep{trenberth2011changes}. Reliable precipitation data are essential for understanding atmospheric processes, evaluating climate models, and supporting early warning systems for extreme events such as floods and droughts \citep{allan2010human, groisman2005trends}. However, the high spatial and temporal variability of precipitation poses significant challenges for data collection, modeling, and forecasting.

Daily precipitation data collected from weather stations across Western and Central Europe \citep{toreti2014gridded} are utilized in this study. These data provide a comprehensive observational record of rainfall patterns, enabling an analysis of spatial and temporal variations in precipitation. The dataset originates from national meteorological agencies and international climate monitoring networks, ensuring high reliability and consistency in measurements.

Daily accumulated precipitation values recorded at multiple weather stations spanning several decades are included, with most locations having records dating back to at least the early 1980s. The stations are part of networks such as the \textit{European Climate Assessment \& Dataset} (ECA\&D) \citep{klein_tank_2002} and the \textit{Global Historical Climatology Network} (GHCN) \citep{menne2012overview}, along with other national meteorological services. The spatial distribution of stations varies, with denser coverage in Western European countries such as France, Germany, Italy, and the Netherlands, while Central European regions, including Poland and the Czech Republic, maintain moderate station density.

Extensive quality control procedures have been applied to ensure data accuracy. Erroneous readings caused by sensor malfunctions or extreme outliers are identified and removed, and missing values are addressed using suitable interpolation techniques. Homogenization procedures are further applied to correct for changes in station locations and measurement methods \citep{venema2012benchmarking}, thereby enhancing robustness for long-term climate studies.

Due to its high temporal resolution, the dataset has been widely used for climate change assessments, hydrological modeling, and studies of extreme weather events \citep{haylock2008european}. In the present work, the potential of Spatio-temporal DeepKriging \citep{NAG2023100773}, implemented in PyTorch, is explored for enhancing both interpolation and forecasting tasks using this dataset.

\section{Analysis}

\begin{figure}[htbp]
    \centering
    \includegraphics[]{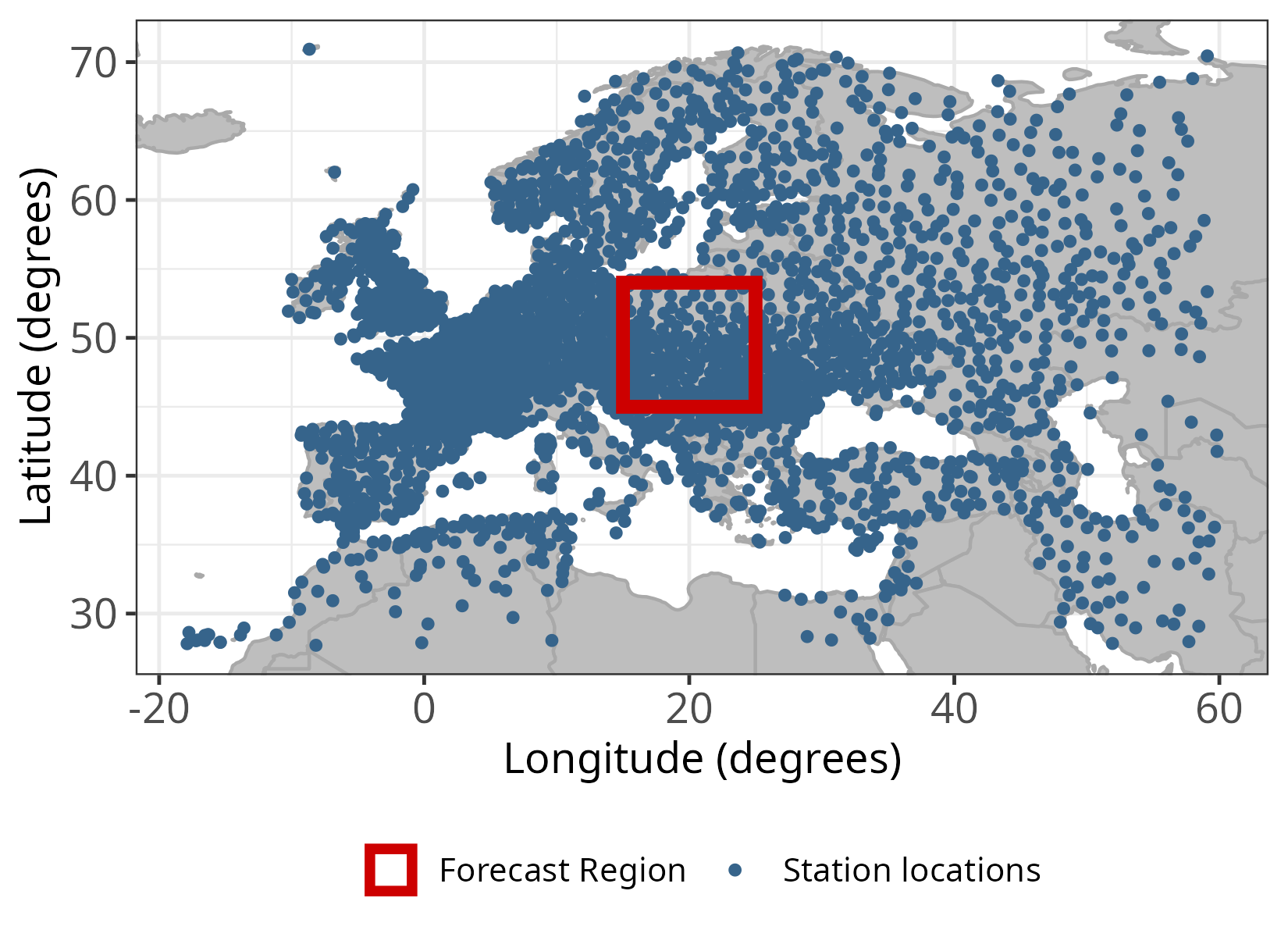}
    \caption{Geographical distribution of weather station locations across Europe. The sub-region marked in red is used for forecasting experiments.}
    \label{fig:precip-europe}
\end{figure}

The dataset consists of daily precipitation measurements collected from weather stations distributed across Europe, spanning the period from January 1, 2014, to December 31, 2024. The spatial distribution of these stations is illustrated in Figure~\ref{fig:precip-europe}, where the sub-region selected for forecasting experiments is highlighted in red.

To reduce short-term variability, a 10-day moving average was applied to the precipitation series. Subsequently, all observations were standardized to achieve consistent scaling across different spatial locations and time periods, thereby improving model stability and comparability.

For the interpolation task, the Spatio-Temporal DeepKriging (STDK) model \citep{NAG2023100773} was applied to generate precipitation predictions at 10-day intervals across approximately 13 million space-time locations within the study domain. The model incorporates a quantile loss function to facilitate uncertainty estimation, with further details on its formulation and integration into the STDK framework provided in \citet{NAG2023100773}. The model incorporates a multi-resolution basis function representation for both spatial and temporal components. Specifically, spatial embeddings utilized four layers of Wendland basis functions with $9^2$, $17^2$, $35^2$, and $73^2$ basis functions at each respective resolution level. The temporal embeddings employed three layers of Gaussian basis functions comprising 50, 350, and 1000 bases with varying ranges to capture temporal variability across multiple scales. Further methodological details regarding these basis function representations can be found in \cite{NAG2023100773}. 

To better align the spatial basis functions with the actual observation network, basis function representations for locations situated over water bodies were removed, thereby emulating the spatial configuration of the station network. In total, the model extracted 1,400 temporal and 6,756 spatial basis functions at each location, capturing both fine-scale local patterns and broader regional structures.

The STDK model consists of a fully connected feedforward neural network trained to map the extracted spatio-temporal basis function representations to the target precipitation values. The network is composed of a total of 10 hidden layers. The first five layers contain 100 nodes each, while the subsequent four layers consist of 50 nodes each. All hidden layers utilize the ReLU activation function \citep{nair2010rectified} to introduce non-linearity and promote efficient training. The final output layer contains three nodes, which directly estimate the predictive median, and the lower and upper bounds corresponding to the 95\% prediction interval. A specialized activation function is applied at the output layer, as described in Equation 7 of \citet{NAG2023100773}, to ensure appropriate scaling and separation of the quantiles during training. The model is trained using the quantile loss function, which enables direct estimation of predictive uncertainty.

Model performance for the interpolation task was assessed using several evaluation metrics, including mean squared prediction error (MSPE), prediction interval coverage probability (PICP), and mean prediction interval width (MPIW) \citep[see][for a detailed discussion on PICP and MPIW]{lakshminarayanan2017simple}. The results are summarized in Table~\ref{tab:metrics}. 

\begin{figure}[htbp]
    \centering
    \resizebox{0.98\textwidth}{!}{
    \begin{tabular}{c ccc}
    \toprule
        \raisebox{10\height}{\centering \textbf{Inputs}} &
         &
         {\includegraphics[]{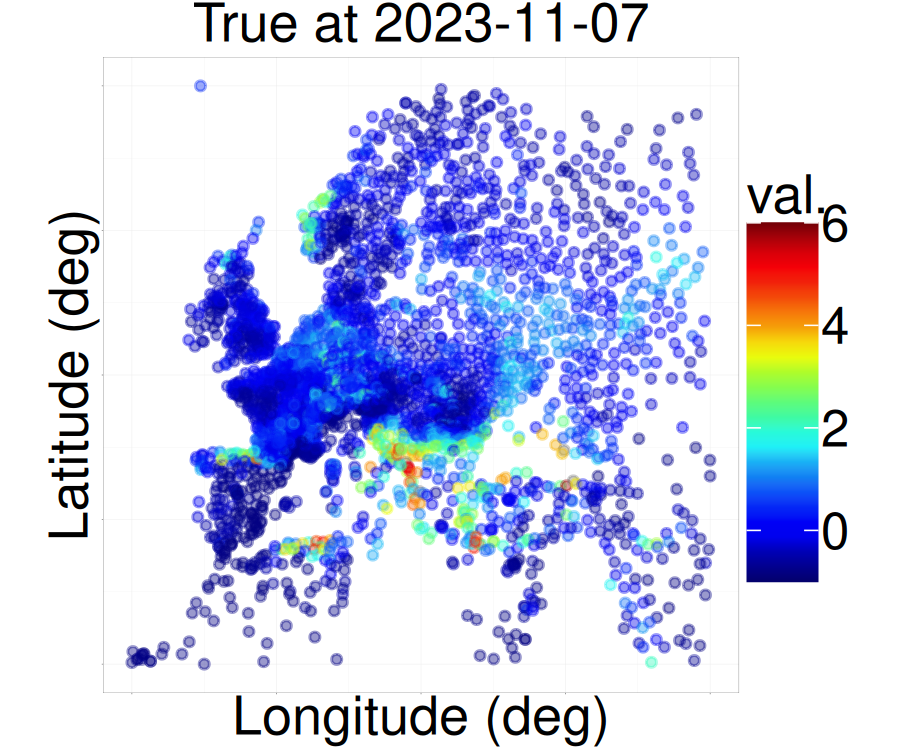}}
        & \\
    \midrule    
        \raisebox{10\height}{\centering \textbf{STDK}} & 
        {\includegraphics[]{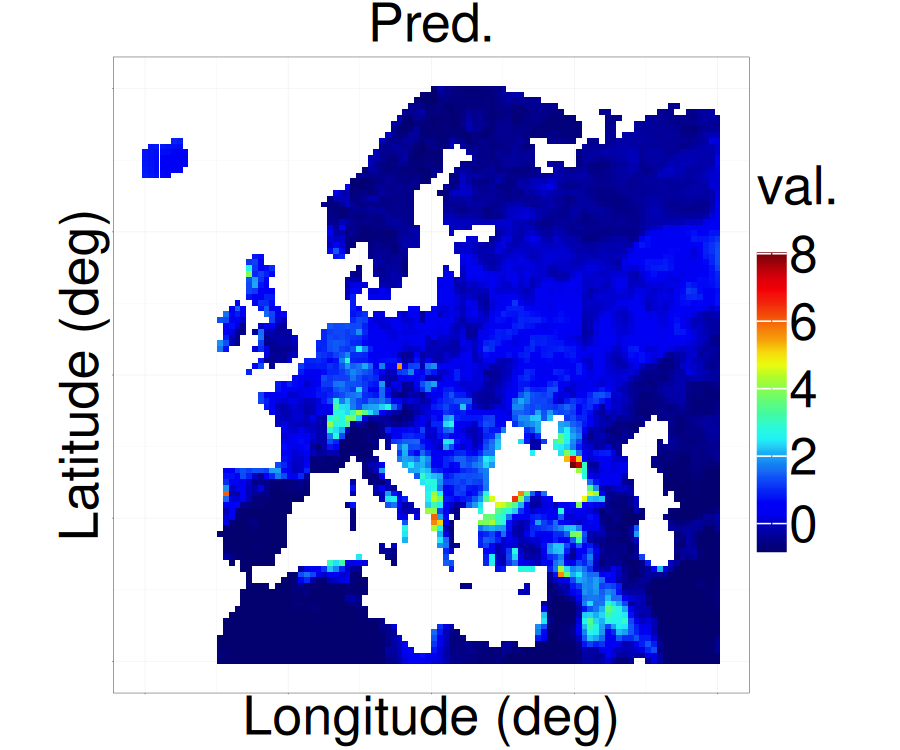}} &
        {\includegraphics[]{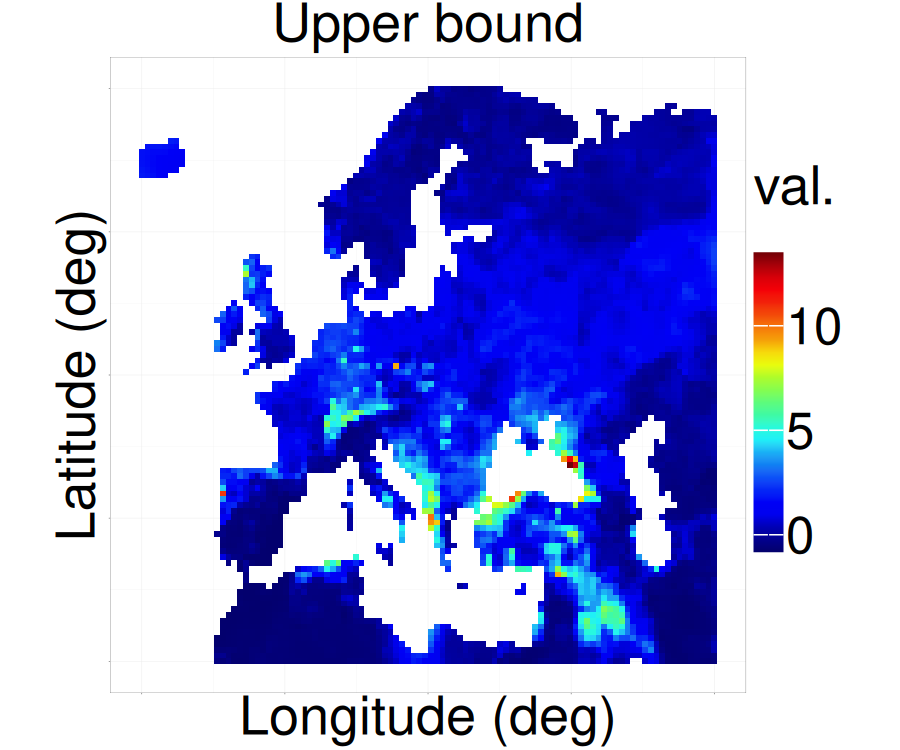}} &
        {\includegraphics[]{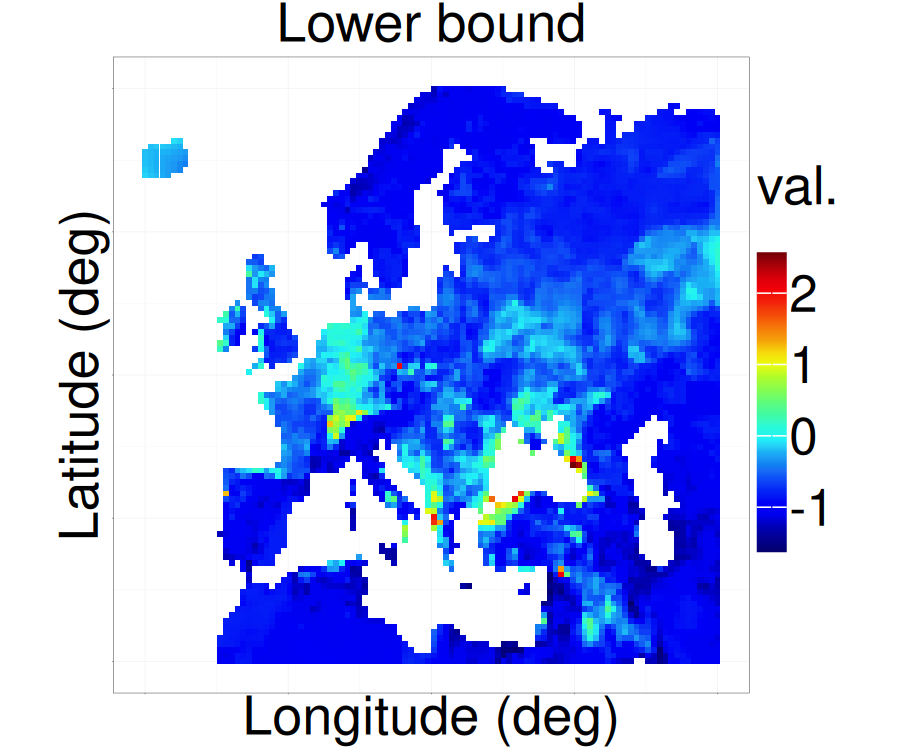}} \\
    \bottomrule     
    \end{tabular}
    }
    \caption{Interpolation results for November 7, 2023. The first column shows the observed precipitation values, while the second, third, and fourth columns display the STDK model's median interpolation, the corresponding 95\% prediction upper bound, and the lower bound, respectively.}
    \label{fig:precip-interp}
\end{figure}

The interpolation results for a representative date are visualized in Figure~\ref{fig:precip-interp}. The true precipitation field, along with the interpolated median predictions and their associated 95\% prediction intervals (upper and lower bounds), are displayed. It is evident from the visualization that the precipitation data exhibits a strong right-skewed distribution, with many areas experiencing no precipitation and relatively few regions recording significant rainfall. This imbalance poses challenges for conventional models; however, the STDK framework successfully identifies the spatial structure of high rainfall areas and provides accurate uncertainty quantification, demonstrating its ability to handle non-Gaussian, highly skewed datasets effectively.

\begin{table}[htbp]
    \centering
    \caption{Performance metrics for the STDK model in the interpolation task.}
    \begin{tabular}{|c|c|}
        \hline
        \textbf{Metric} & \textbf{Value} \\
        \hline
        MSPE  & 0.017 \\
        \hline
        PICP  & 0.955 \\
        \hline
        MPIW  & 0.245 \\
        \hline
    \end{tabular}
    \label{tab:metrics}
\end{table}

In addition to interpolation, forecasting experiments were conducted to evaluate the model's predictive capabilities. For this purpose, a $64 \times 64$ spatial grid was generated within the sub-region marked in red in Figure~\ref{fig:precip-europe}. The STDK model was first used to generate interpolated images over this forecasting domain at each time point. These interpolated images were then used to train forecasting models. Each forecasting model received a sequence of three consecutive interpolated images as input to predict the precipitation field at the sixth time step. A total of 3,900 training image sequences were constructed for model training.

\begin{figure}[htbp]
    \centering
    \resizebox{0.98\textwidth}{!}{
    \begin{tabular}{c ccc}
    \toprule
        \raisebox{10\height}{\centering \textbf{Inputs}} & &
        {\includegraphics[]{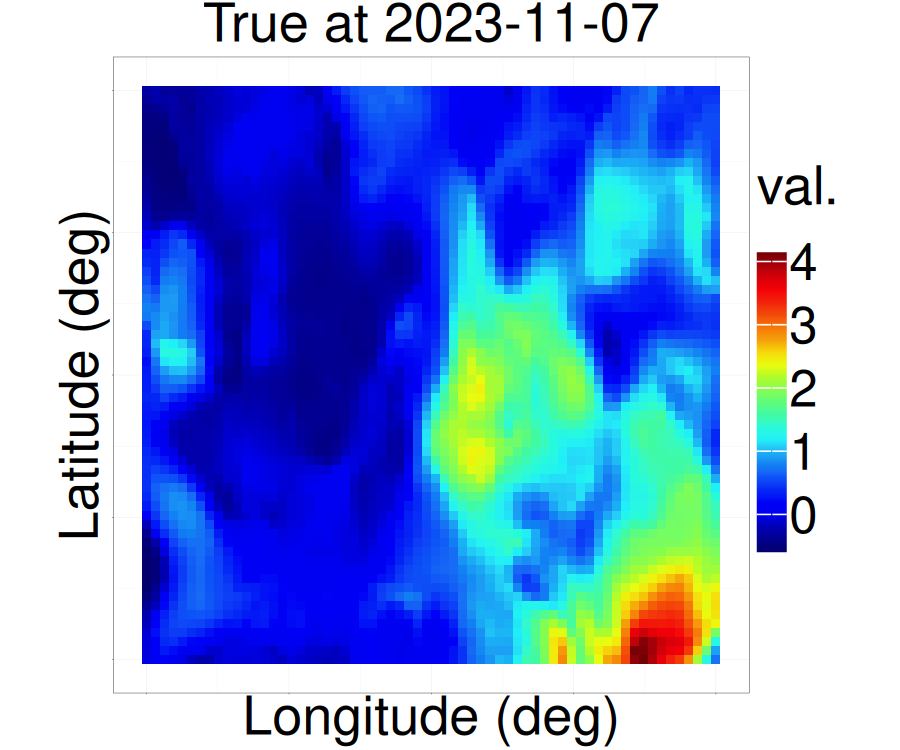}}
         & \\
    \midrule
        \raisebox{10\height}{\centering \textbf{ConvLSTM}} & 
        {\includegraphics[]{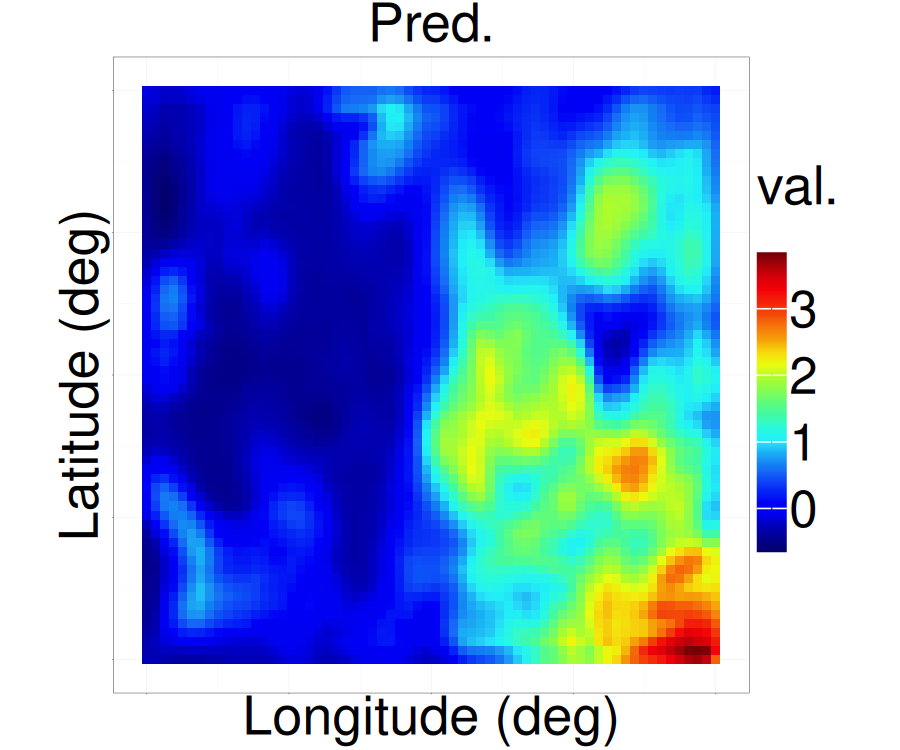}} &
        {\includegraphics[]{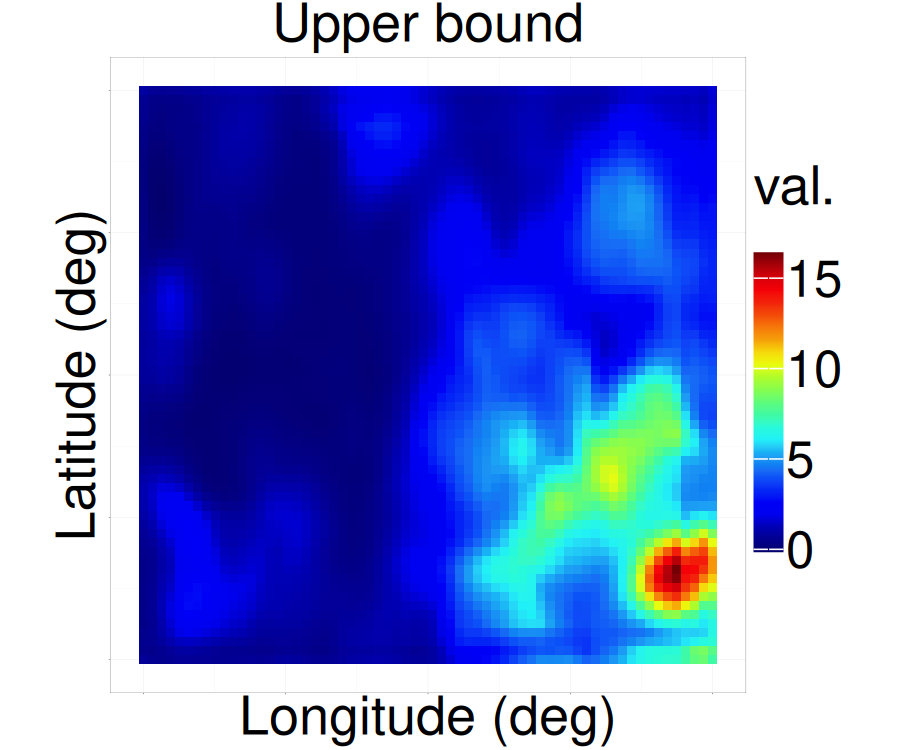}} &
        {\includegraphics[]{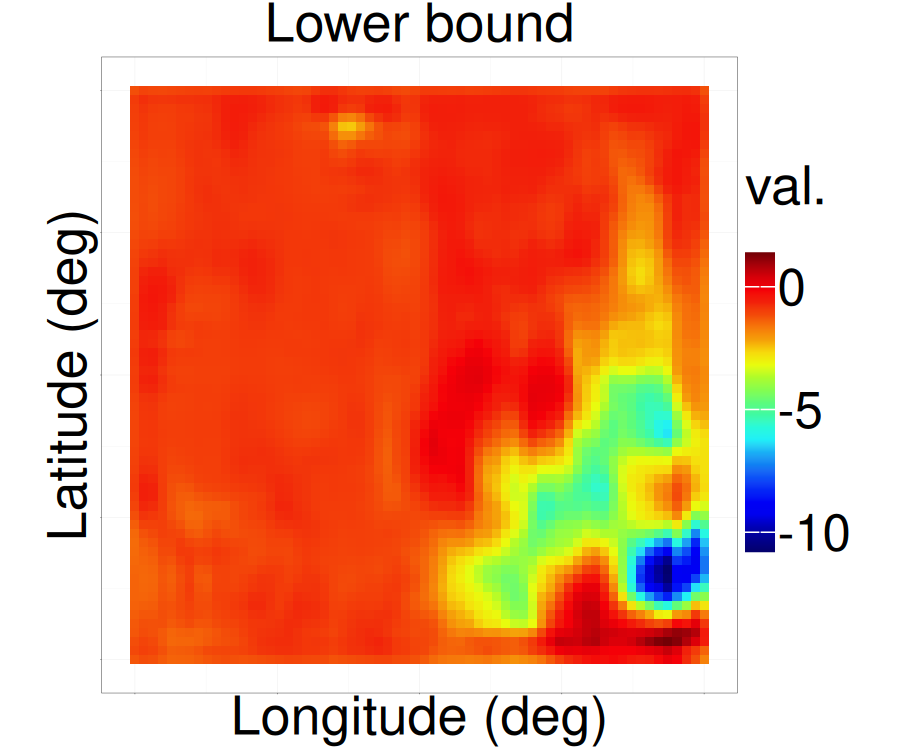}} \\
    \midrule    
        \raisebox{10\height}{\centering \textbf{STDK}} & 
        {\includegraphics[]{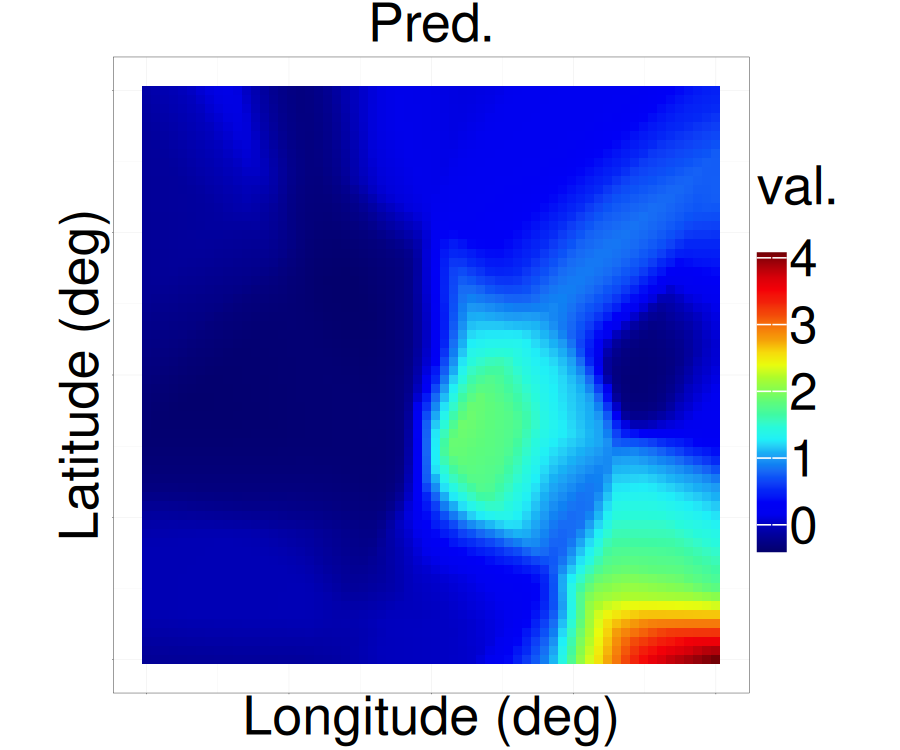}} &
        {\includegraphics[]{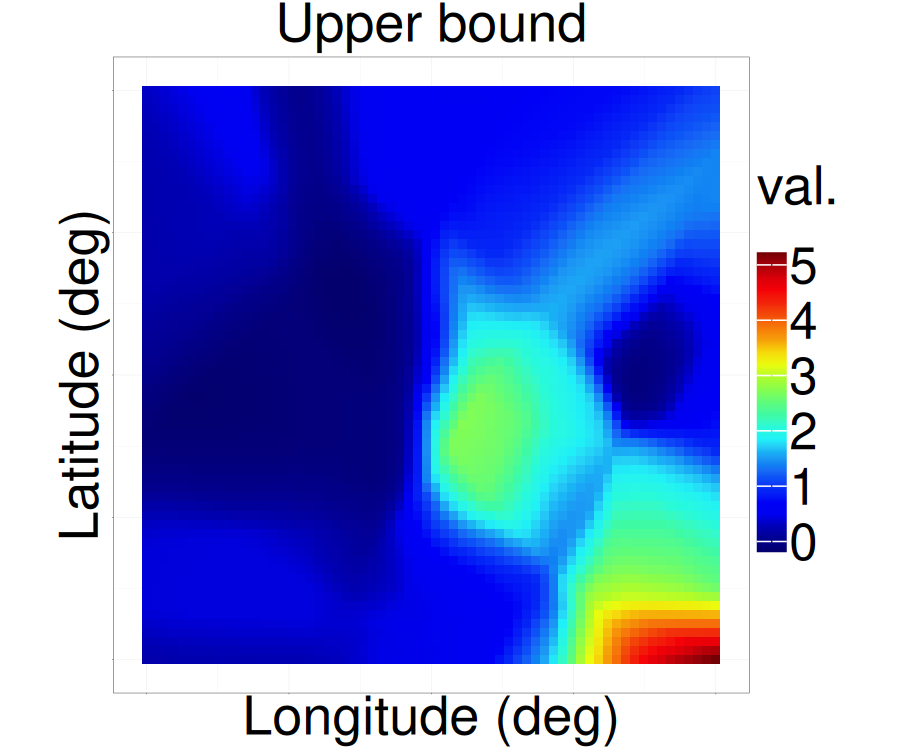}} &
        {\includegraphics[]{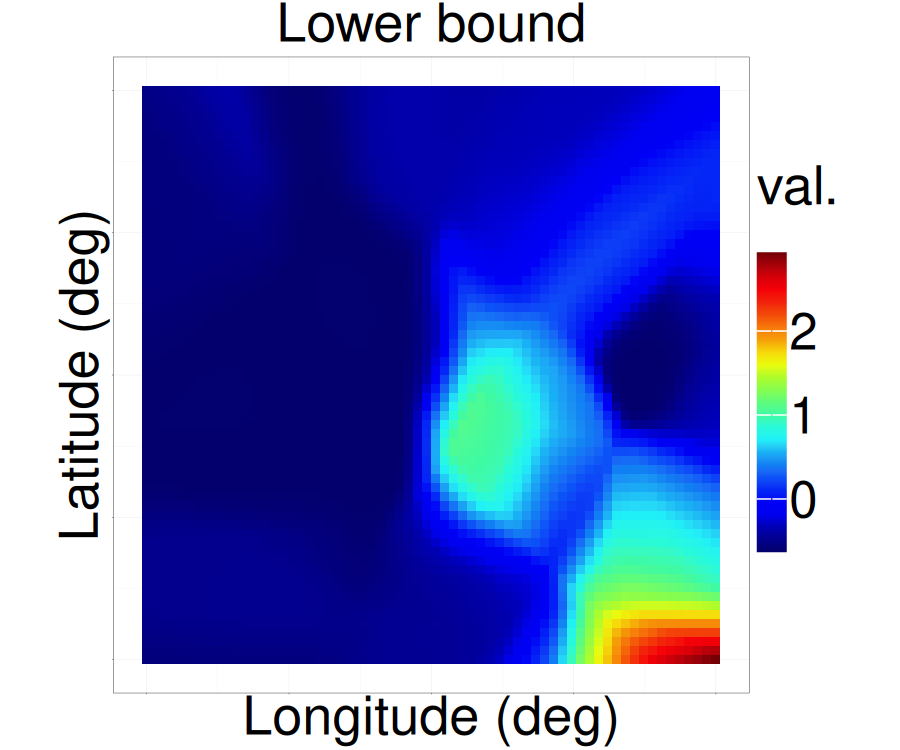}} \\
    \bottomrule     
    \end{tabular}
    }
    \caption{Forecasting results for November 7, 2023. The first row displays the true precipitation field at the target time point. The second and third rows present results for ConvLSTM and STDK models, respectively. For each model, columns show the predicted precipitation field, the corresponding 95\% prediction upper bound, and lower bound.}
    \label{fig:precip-forecast}
\end{figure}

The forecasting performance of the ConvLSTM and STDK models was compared in Table~\ref{tab:precip}, with both models trained using a quantile loss function to enable uncertainty estimation.
Figure~\ref{fig:precip-forecast} presents a visual diagnostic comparison of both models for a representative test case. For each model, the predicted precipitation field, along with its 95\% upper and lower prediction bounds, are displayed. The comparison demonstrates the effectiveness of STDK in generating sharper and more accurate forecasts while properly quantifying the associated uncertainties.
\begin{table}[h]
    \centering
    \begin{tabular}{|lccc|}
        \hline
        & MSPE & PICP & MPIW \\
        \hline
        ConvLSTM & 0.060 & 0.94 & 0.888 \\
        STDK     & 0.101 & 0.95 & 0.783 \\
        \hline
    \end{tabular}
    \caption{Performance metrics of comparing models for forecasting.}
    \label{tab:precip}
\end{table}

\section{Conclusion}
In this study, a PyTorch-based implementation of the Spatio-Temporal DeepKriging (STDK) framework has been developed and applied to the problem of precipitation data analysis over Europe. Leveraging a multi-resolution basis function representation, the model successfully captures complex spatio-temporal dependencies inherent in the precipitation dataset, which exhibits substantial skewness and sparsity across space and time. The interpolation results demonstrate that STDK can accurately reconstruct precipitation fields, even in the presence of irregular station networks and highly localized precipitation events. The model's ability to deliver reliable uncertainty quantification through well-calibrated prediction intervals further highlights its robustness for practical environmental applications. The forecasting experiments additionally demonstrate the capacity of the quantile-based ConvLSTM, particularly in terms of providing sharper and more stable probabilistic forecasts.

Beyond the specific application to precipitation data, the methodological framework presented here offers a flexible and scalable approach for handling irregularly spaced spatio-temporal datasets, which are common in many environmental and geophysical domains. The modular implementation in PyTorch not only ensures reproducibility but also allows for straightforward customization and extension to different datasets. As climate and environmental monitoring continue to generate increasingly complex datasets, models like STDK provide an important tool for researchers and policymakers seeking accurate predictions and interpretable uncertainty estimates. 
\\
\\

\bibliographystyle{apalike}
\bibliography{ref}

\end{document}